\documentclass[runningheads]{llncs}
\usepackage{graphicx}
\usepackage{amsmath,amssymb} 
\usepackage{color}
\usepackage{bm}
\usepackage{multirow}
\usepackage{diagbox}
\usepackage{array}
\usepackage{booktabs}
\usepackage{nth}
\usepackage{nicefrac}
\usepackage{relsize}
\usepackage{xspace} 
\usepackage{bm}
\usepackage{enumerate}
\usepackage{subcaption}
\usepackage[table]{xcolor}
\usepackage{enumitem}
\usepackage{array}
\usepackage[unicode,colorlinks=true,pagebackref=false]{hyperref}
\usepackage[width=122mm,left=12mm,paperwidth=146mm,height=193mm,top=12mm,paperheight=217mm]{geometry}
\newcolumntype{P}[1]{>{\centering\arraybackslash}p{#1}}

\newcommand{\etal}{\textit{et al. }}
\newcommand{\eg}{e.g.}
\newcommand{\ie}{i.e.}
\begin{document}
\pagestyle{headings}
\mainmatter

\title{Deep Imbalanced Attribute Classification using Visual Attention Aggregation}
\titlerunning{Deep Imbalanced Attribute Classification using Attention Aggregation}
\author{Nikolaos Sarafianos, Xiang Xu, and Ioannis A. Kakadiaris}
\institute{Computational Biomedicine Lab\\
	University of Houston\\
	\email{ \{nsarafia, xxu18, ikakadia\}@central.uh.edu}
}
\maketitle
\begin{abstract}
	For many computer vision applications, such as image description and human identification, recognizing the visual attributes of humans is an essential yet challenging problem. Its challenges originate from its multi-label nature, the large underlying class imbalance and the lack of spatial annotations. Existing methods follow either a computer vision approach while failing to account for class imbalance, or explore machine learning solutions, which disregard the spatial and semantic relations that exist in the images. With that in mind, we propose an effective method that extracts and aggregates visual attention masks at different scales. We introduce a loss function to handle class imbalance both at class and at an instance level and further demonstrate that penalizing attention masks with high prediction variance accounts for the weak supervision of the attention mechanism. By identifying and addressing these challenges, we achieve state-of-the-art results with a simple attention mechanism in both PETA and WIDER-Attribute datasets without additional context or side information. 
\end{abstract}
\section{Introduction}
We set out to develop a method that, given an image of a human, predicts its visual attributes. We posed the following questions: (i) what are the challenges of this problem? (ii) what have other people done? and (iii) how should a simple yet effective solution to this problem look like? Human attributes are imbalanced in nature. Bald individuals with a mustache wearing glasses are 14 to 43 times less likely to appear in the CelebA dataset~\cite{liu2015faceattributes} compared to people without these characteristics. Large-scale imbalanced datasets can lead to biased models, optimized to favor the majority classes while failing to identify the subtle discriminant features that are required to recognize the under-represented classes. Setting the class imbalance aside, an additional challenge is identifying which areas in the image provide class-discriminant information. Giving emphasis to the upper part of an image, where the face is located, for attributes such as ``glasses'' and to the bottom part for attributes such as ``long pants'' can increase the recognition performance as well as the interpretability of our models~\cite{olah2018the}. This challenge is usually addressed using visual attention techniques that output saliency maps. However, in the human attribute estimation domain, attention ground-truth annotations are not available to learn such spatial attributions. 

Learning from imbalanced data is a well-studied problem in machine learning and computer vision. Traditional solutions include over-sampling the minority classes~\cite{chawla2002smote,maciejewski2011local} or under-sampling the majority classes~\cite{drummond2003c4} to compensate for the imbalanced class ratio and cost-sensitive learning~\cite{khan2017cost} where classification errors are penalized differently. Such approaches have been extensively used in the past but they suffer from some limitations. For example, over-sampling introduces redundant information making the models prone to over-fitting, whereas under-sampling may remove valuable discriminative information. Recent works with deep convolutional neural networks~\cite{huang2016learning,huang2018deep,dong2017class} introduced a sampling procedure of triplets, quintuplets or clusters of samples that satisfy some properties in the feature-space and used them to regularize their models. However, sampling triplets is a computationally expensive procedure and the characteristics of the triplets in a batch-mode setup might vary significantly. 

Modern visual attribute classification techniques rely either on contextual information~\cite{li2016human,gkioxari2015actions}, side information~\cite{sarfraz2017deep}, curriculum learning strategies~\cite{sarafianos2018curriculum} or visual attention mechanisms~\cite{zhu2017learning} to accomplish their task. Although context and side information can increase the recognition accuracy, we believe that a simple solution should not rely on those. We argue that a solution to the deep imbalanced attribute classification problem should: (i) extract discriminative information, (ii) leverage visual information that is specific for each attribute, and (iii) handle class imbalance. Since, to the best of our knowledge, there is no method available with such characteristics, we developed an approach that uses (i) a pre-trained network for feature extraction, (ii) a weakly-supervised visual attention mechanism at multiple scales for attribute specific information, and (iii) a loss function that handles class imbalance and focuses on hard and uncertain samples. By simplifying the problem and addressing each one of its challenges, we were able to achieve state-of-the-art results in both WIDER-Attribute~\cite{li2016human} and PETA~\cite{deng2014pedestrian} datasets, which are the most widely used in this domain. 

\begin{figure*}[t]
	\centering
	\includegraphics[width=0.99\linewidth]{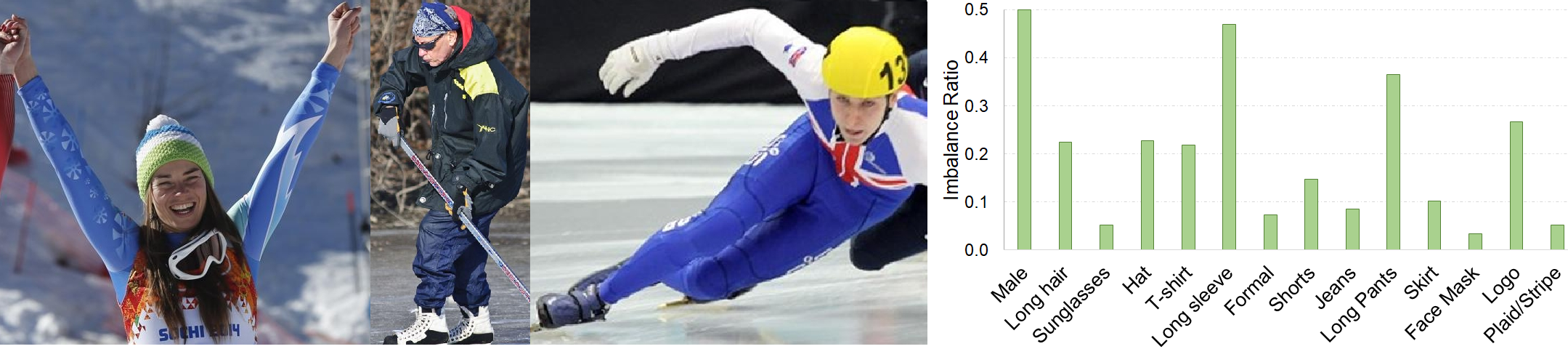}
	\caption{Visual attribute classification challenges from left to right: (i) the face mask is under the head, (ii) are there sunglasses in the image? (iii) extreme pose variation, and (iv) large class imbalance.}
	\label{fig:challenges}
\end{figure*}

In the deep learning era, most models are overly-complicated for what they aspire to achieve. Carefully developed, well established, accurate baselines are essential to measure our progress over time. Towards this direction, there have been a few works recently with well-performing yet simple baseline approaches in the fields of 3D human pose estimation~\cite{martinez2017simple}, image classification~\cite{chan2015pcanet}, and person re-identification~\cite{zheng2017unlabeled}. Our main contribution is the design and analysis of an end-to-end neural-network architecture that can be easily reproduced, is easy to train and achieves state-of-the-art visual attribute classification results. This performance improvement originates from extracting and aggregating visual attention masks at different scales as well as establishing a loss function for imbalanced attributes as well as hard or uncertain samples. Through experiments, ablation studies and qualitative results we demonstrate that:
\begin{itemize}
	\item A simple visual attention mechanism with only attribute-level supervision (no ground-truth attention masks) can improve the classification performance by guiding the network to focus its resources to those spatial parts that contain information relevant to the input image.
	\item Extracting visual attention masks from more than one stage of the network and aggregating the information at a score-level enables the model to learn more discriminant feature representations. 
	\item Accounting for class imbalance is essential during learning from large datasets. While assigning prior class weights can alleviate part of this problem, we observed that a weighted-variant of the focal loss works consistently better by handling imbalanced classes and at the same time focusing on hard examples. 
	\item Due to the lack of strong supervision, the attention masks result in attribute predictions with high variance across subsequent epochs. To prevent this from destabilizing training and degrading the performance we introduce an attention loss function, which penalizes predictions that originate from attention masks with high prediction variance. 
\end{itemize}
Since this work aspires to serve as a bar in the visual attribute classification domain that future works may improve upon, we identify some sources of error that still prevail, and point out future research directions to address them that require further exploration. 

\section{Related Work}
\noindent~\textbf{Visual Attributes}: When we are interested in providing a description of an object or a human, we tend to rely on visual attributes to accomplish this task. From early works~\cite{ferrari2007learning,kumar,chen2012describing} to more recent ones \cite{li2016human,sarafianos2017curriculum,gkioxari2015actions,sarfraz2017deep,zhu2017learning,zhang2014panda} visual attributes have been studied extensively in computer vision. Due to its commercial applications and the abundance of available data, the clothing domain has received significant attention recently with methods ranging from transfer learning and domain adaptation~\cite{dong2017multi,sarafianos2017adaptive,chen2015deep} to retrieval~\cite{liu2016deepfashion} and forecasting~\cite{al2017fashion}. Some works rely on contextual information \cite{li2016human,gkioxari2015actions}, or leverage side information (\eg, viewpoint) to improve the recognition performance~\cite{sarfraz2017deep}. Others \cite{zhu2016multi}, assume the existence of a predefined connection between parts and attributes (\eg, hats are usually above the head and in the upper 20\% of the image) which does not always hold true as depicted in Figure~\ref{fig:challenges}. Zhu \etal~\cite{zhu2017learning} proposed to learn spatial regularizations using an attention mechanism on a final ResNet~\cite{resNets2016} representation. Their attention module outputs an attention tensor per attribute which is then fed to a multi-label classification sub-network.  However, none of the aforementioned approaches consider the class imbalance that exists in such datasets, which prevents them from accurately recognizing under-represented attributes such as wearing sunglasses. 

\noindent~\textbf{Visual Attention}: Visual attention can be interpreted as a mechanism of guiding the network to focus its resources on those spatial parts that contain information relevant to the input image. In computer vision applications, visual attribution is usually implemented as a gating function represented with a sigmoid activation or a spatial softmax and is placed on top of one or more convolutional layers with small kernels extracting high-level information. Several interesting works have appeared recently that demonstrate the efficiency of visual attention \cite{zhu2017learning,wang2017face,liu2016fully,li2018harmonious,wang2017residual,hu2017squeeze,chu2017multi,chen2017order}. For example, the harmonious attention of Li \etal~\cite{li2018harmonious} consists of four subparts that extract hard-regional attention, soft-spatial, and channel attention to perform person re-identification. Deciding where to place the attention mechanism in the network is a topic of active research with several single-scale and multi-scale attention techniques in the literature. Das \etal~\cite{das2017human}, opted for a single attention module, whereas others~\cite{wang2017face,chu2017multi,rodríguez2018a} extract saliency heatmaps at multiple-scales to build richer feature representations. 

\noindent~\textbf{Deep Imbalanced Classification}: Two works that address this problem in an attribute classification framework are the large margin local embedding (LMLE) method~\cite{huang2016learning} and the class rectification loss (CRL)~\cite{dong2017class}. In LMLE, quintuplets were sampled that preserve locality across clusters and discrimination between classes and a new loss was introduced. Dong~\etal~\cite{dong2017class} demonstrated that a careful hard mining of triplets within the batch acts as an effective regularization which improves the recognition performance of imbalanced attributes. However, LMLE is prohibitively computationally expensive as it comprises an alternating scheme for cluster refinement and classification. In a follow-up work~\cite{huang2018deep} the authors address this limitation by replacing the quintuplets with clusters. CRL on the other hand, samples triplets within the batch, complicating the training process significantly, as the convergence and the performance heavily rely on the triplet selection. In addition, CRL adds a fully-connected layer for each attribute before the final classification layer, which increases the number of parameters that need to be learned. Both methods approach class imbalance purely as a machine learning problem without focusing on the visual traits of the images that correspond to these attributes. Class imbalance arises also in detection problems~\cite{wang2017face,lin2017focal}, where the foreground object (or face) covers a small part of the image. A simple yet very effective solution is focal loss~\cite{lin2017focal}, which uses a weighting scheme at an instance-level within the batch to penalize hard misclassified samples and assign near-zero weights to easily classified samples. 

\section{Methodology}
\begin{figure*}[t]
	\centering
	\includegraphics[width=0.99\linewidth]{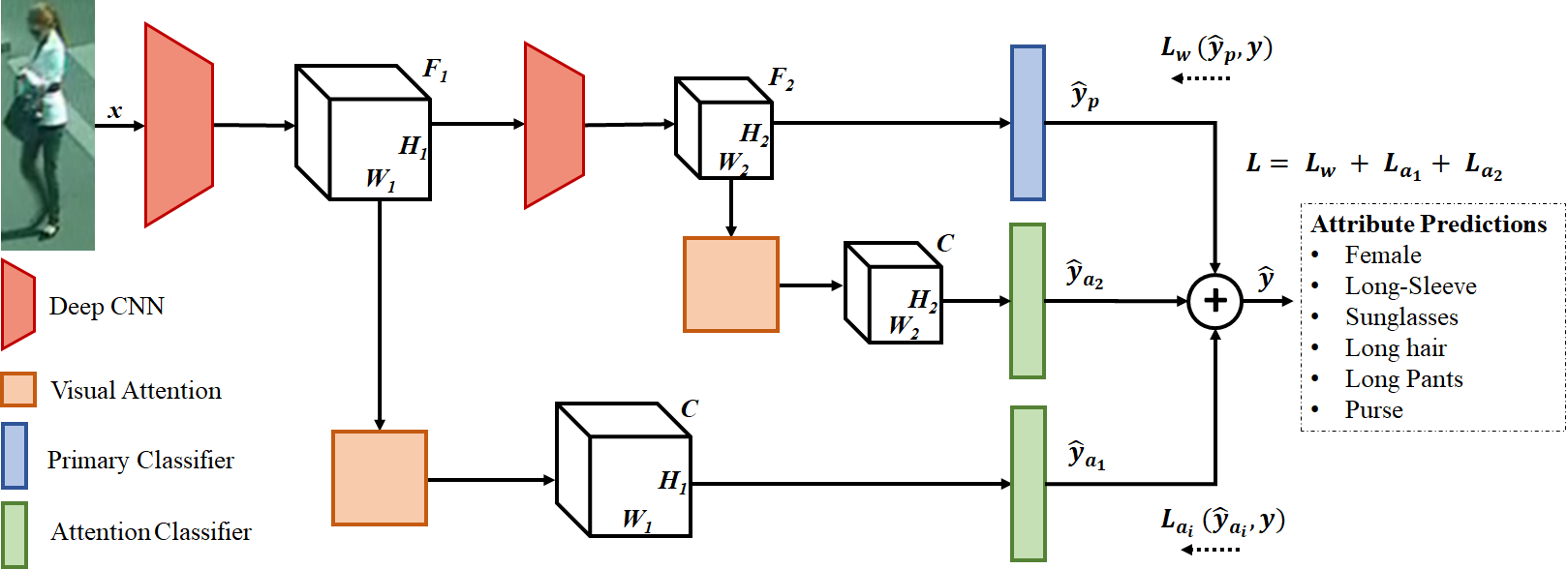}
	\caption{Given an image of a human we aspire to predict \(C\) visual attributes. Visual attention mechanisms are placed at two different levels of the network to identify spatial information that is relevant to each attribute with only attribute-level supervision. The predictions from the attention and the primary classifiers are aggregated at a score level and the whole network is trained end-to-end with two loss functions: \(\mathcal{L}_{w}\) that handles class imbalance and hard samples and \(\mathcal{L}_{a}\) which penalizes attention masks with high prediction variance.}
	\label{fig:diag}
\end{figure*}


\subsection{Multi-scale Visual Attention and Aggregation}
\begin{figure}[t]
	\centering
	\begin{subfigure}[t]{0.51\textwidth}
		\includegraphics[width=\textwidth]{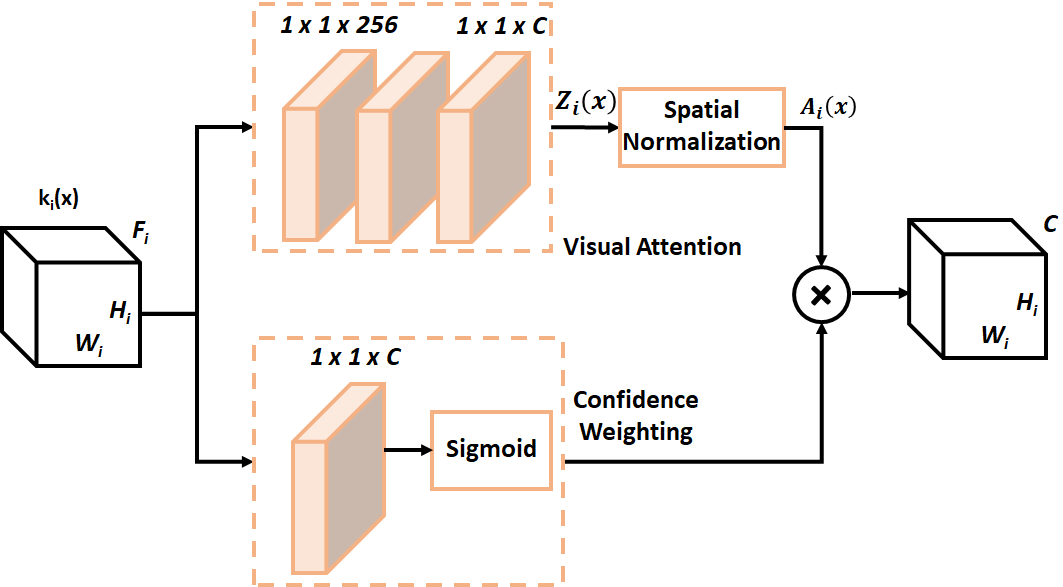}
	\end{subfigure}
	\begin{subfigure}[t]{0.47\textwidth}
		\includegraphics[width=\textwidth]{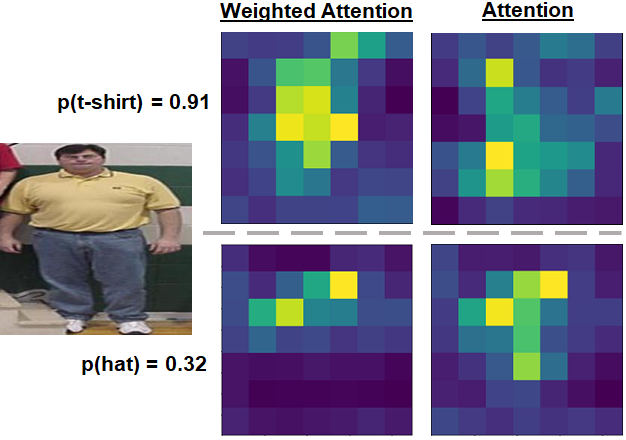}
	\end{subfigure}
	\caption{Our attention mechanism (upper-left) maps feature representations of spatial resolution \(H_i\times W_i\) and \(F_i\) channels to \(C\) channels (one for each attribute) with the same size which are then spatially normalized to force the model to focus its resources to the most relevant region of the image. The attention masks are weighted by attribute confidences (lower-left) which as we demonstrate on the right, apply larger weights to the attribute-corresponding areas. For example, more emphasis is given in the middle-upper part when looking for a t-shirt and to the upper part of the image when looking for a hat (even when it is not there).}
	\label{fig:att}
\end{figure}

Given an image of a human our goal is to predict its visual attributes. Specifically, our input consists of an image \(x\) along with its corresponding labels \(y=[y^1, y^2,\dots,y^C]^T \) where \(C\) is the total number of attributes and \(y^c\) a binary label that indicates the presence or absence of a particular attribute in the image. In this work, we experimented with both ResNets~\cite{resNets2016} and DenseNets~\cite{huang2017densely} as backbone architectures and thus, we opted for the representations after the third and the fourth stage/block of layers. The concept of extracting attention information can be expanded to more spatial resolutions/scales besides two at the expense of learning additional parameters. We will thus refer to the first part of the networks (up to stage/block three) as \(\phi_1(\cdot)\) and to the part from there and until the classifier as \(\phi_2(\cdot)\). In our primary network, which unless otherwise specified is a ResNet-101 architecture (deep CNN module in Figure~\ref{fig:diag}), given an image \(x\), we obtain three-dimensional feature representations:
\begin{equation}
\begin{gathered}
k_1(x) = \phi_1(x),  \; \; k_1(x)\in \mathcal{R}^{H_1 \times W_{1} \times F_{1}}, \\ k_2(x) = \phi_2(k_1(x)),  \; \; k_2(x)\in \mathcal{R}^{H_2 \times W_{2} \times F_{2}}.
\end{gathered}
\end{equation}
For \(224\times 224\) images the attention mechanism is placed on features of channel size \(F_i\) equal to \(1,024\) and \(2,048\) with spatial resolutions \(H_i\times W_i\) equal to \(14\times 14\) and \(7\times 7\), respectively. Finally, the classifier of the primary network outputs logits \(\hat{y}_{p}(x) = W_{p}k_2(x)+b_{p}\) where \( (W_{p}, b_{p})\) are the parameters of the classification layer. 

With simplicity in mind, our attention mechanism, depicted in Figure~\ref{fig:att}, consists of three stacked convolutional layers (along with batch-normalization and ReLU) with a kernel size equal to one. Due to the multi-label nature of the problem, the last convolutional layer maps the channels to the \(C\) number of classes (\ie, attributes). This is different than most attention works (with one label per image) that extract saliency maps of the same spatial/channel size of the given feature representation. The attribute-specific attention maps \(z_{h,w}^c\) are then spatially normalized to \(a_{h,w}^c\)using a spatial softmax operation:
\begin{equation}\label{eq:spat}
a_{h,w}^c = \frac{\exp(z_{h,w}^c)}{\sum_{h,w} \exp(z_{h,w}^c)},
\end{equation}
where \(h,w\) correspond to the height and width dimension and \(c\) to the corresponding attribute label. The spatial softmax operation results in attention masks with the property \( \sum_{h,w} a_{h,w}^c = 1\) for each attribute \(c\) and is used to force the model to focus its resources to the most relevant region of the image. We will refer to the attention mechanism comprising the three convolutional layers as \(\mathcal{A}\) and thus, for each spatial resolution \(i\) we first obtain unnormalized attentions \(Z_i(x) = \mathcal{A}(k_i(x))\), which are then spatially normalized using Eq. (\ref{eq:spat}) resulting in normalized attention masks \(A_i(x)\). 

Following the work of Zhu \etal~\cite{zhu2017learning}, we concurrently pass the feature representations to a single convolutional layer with \(C\) channels (same as the number of classes) followed by a sigmoid function. The role of this branch is to assign weights to the attention maps based on label confidences and avoid learning from the attention masks when the label is absent. The weighted attention maps reflect both attribute information at different spatial locations and label confidences. We observed in our experiments that this confidence-weighting branch boosts the performance by a small amount and helps the attention mechanism learn better saliency heatmaps (Figure~\ref{fig:att} right). 

Combining the output saliency masks from different scales can be done either at a prediction level (\ie, averaging the logits) or at a feature level~\cite{wang2018deep}. However, aggregating the attention masks at a feature level provided consistently inferior performance. We believe that this is because the two attention mechanisms extract masks that give emphasis to different spatial regions which, when added together, fail to provide the classifier with attribute-discriminative information. Thus, we opted for the former approach and fed each confidence-weighted attention mask to a classifier to obtain logits \(\hat{y}_{a_i}\) of the attention module \(i\). The final attribute predictions of dimensionality \(1\times C\) for an image \(x\) are then defined as \(\hat{y} = \nicefrac{(\hat{y}_{p} + \hat{y}_{a_{1}} + \hat{y}_{a_{2}})}{3}\). 

\subsection{Deep Imbalanced Classification}
Using the output predictions of the primary model \(\hat{y}_{p}\) which have the same dimensionality \(1\times C\) (\ie, one for each attribute), a straight-forward approach adopted by Zhu \etal~\cite{zhu2017learning} is to train the whole network using the binary cross-entropy loss \(\mathcal{L}_b\) as:
\begin{equation}
\mathcal{L}_b(\hat{y}_{p}, y) = - \sum_{c=1}^{C} \log(\sigma(\hat{y}_{p}^c)) y^c + \log(1 -\sigma(\hat{y}_{p}^c)) (1-y^c),
\end{equation}
where \((\hat{y}_{p}^c, y^c)\) correspond to the logit and ground-truth labels for attribute \(c\), and \(\sigma(\cdot) \) is the sigmoid activation function. However, such a loss function ignores completely the class imbalance. Aiming to alleviate this problem both at a class- and at an instance-level, we propose to use for our primary model a weighted-variant of the focal loss~\cite{lin2017focal} defined as:
\begin{small}
	\begin{multline}
	\mathcal{L}_{w}(\hat{y}_{p}, y) = - \sum_{c=1}^{C} w_c \Big( \big(1-\sigma(\hat{y}_{p}^c)\big)^\gamma \log\left(\sigma(\hat{y}_{p}^c)\right)y^c \; + \; \sigma(\hat{y}_{p}^c)^\gamma \log(1 -\sigma(\hat{y}_{p}^c))(1-y^c) \Big),
	\end{multline}
\end{small}

\noindent where \(\gamma\) is a hyper-parameter (set to \(0.5\)), which controls the instance-level weighting based on the current prediction giving emphasis to the hard misclassified samples, and \(w_c = e^{-a_c}\), where \(a_c\) the prior class distribution of the \(c^{th}\) attribute as in \cite{sarfraz2017deep}. 

Unlike the face attention networks~\cite{wang2017face}, which learn the attention masks based on ground-truth facial bounding boxes, in the human attribute domain such information is not available. This means that the attention masks will be learned based on attribute-level supervisions \(y\). The attention masks of dimensionality \(H_i \times W_{i} \times F_{i}\) are fed to a classifier which outputs logits \(\hat{y}_{a_i}\) for each spatial resolution \(i\). To account for the weak supervision of the attention network, we decided to focus on the attention masks with high prediction variance. Similar to the work of Chang \etal~\cite{chang2017active}, after some burn-in epochs in which \(\mathcal{L}_b\) is used, we start collecting the history \(H\) of the predictions \(p_H(y_s|x_s)\) for the \(s^{th}\) sample and compute the standard deviation across time for each sample within the batch:
\begin{equation}
\small
\widehat{std}_s(H) = \sqrt{\widehat{var}\big(p_{H^{t-1}}(y_s|x_s)\big) + \frac{\widehat{var}\big(p_{H^{t-1}}(y_s|x_s)\big)^2}{|H_s^{t-1}|-1}},
\end{equation}
where \(t\) corresponds to the current epoch, \(\widehat{var}\) to the prediction variance estimated in history \(H^{t-1}\) and \(|H_s^{t-1}|\) the number of stored prediction probabilities. The loss for the attention-masks at level \(i\) with attribute-level supervision for each sample \(s\) is defined as: 
\begin{equation}
\small
\mathcal{L}_{a_i}(\hat{y}_{a_i}, y) = \big(1 + \widehat{std_s}(H)\big)\mathcal{L}_b(\hat{y}_{a_i}, y) \;.
\end{equation}
Attention mask predictions with high standard deviation across time will be given higher weights in order to guide the network to learn those uncertain samples. Note that for memory reasons, our history comprises only the last five epochs and not the entire history of predictions. We believe that such a scheme makes intuitively more sense in a weakly-supervised application rather than the fully-supervised scenarios (such as MNIST or CIFAR) in the original paper \cite{chang2017active}. Finally, the total loss that is used to train our network end-to-end (the primary network and the two attention modules) is defined as:
\begin{equation}
\small
\mathcal{L} = \mathcal{L}_{w} + \mathcal{L}_{a_{1}} + \mathcal{L}_{a_{2}}, 
\end{equation}
where \(\mathcal{L}_{a_1}\) is applied to the first attention module that extracts saliency maps of spatial resolution \(14\times 14\), and \(\mathcal{L}_{a_2}\) is similarly applied to the second attention module after the fourth stage of the primary network with spatial resolution of \(7\times 7\).	Disentangling the two loss functions enables us to focus on different types of challenges separately. The weighted focal loss \(\mathcal{L}_{w}\), handles the prior class imbalance per attribute using the weight \(w_c\) and at the same time focuses on hard misclassified positive samples via the instance-level weights of the focal loss. The attention loss \(\mathcal{L}_{a}\) penalizes predictions that originate from attention masks with high prediction variance. 

\section{Experiments}
To assess our method, we performed experiments and ablation studies on the publicly available WIDER-Attribute~\cite{li2016human} and PETA~\cite{deng2014pedestrian} datasets, which are the most widely used in this domain. The training details for both datasets are provided in the supplementary material. 

\setlength{\tabcolsep}{.07cm}
\setlength{\textfloatsep}{0.5cm}
\begin{table}[t]
	\centering
	\caption{Evaluation of the proposed approach against nine different methods. The asterisk next to SRN indicates that it is our re-implementation due to the fact that the validation set was included in the original work which is not the case for the rest of the methods.}
	\label{tab:WIDER}
	\scriptsize
	\begin{tabular}{lcccccccccccccc|c}
		\toprule
		\textbf{Method} & \rotatebox[origin=c]{90}{Male} & \rotatebox[origin=c]{90}{Long hair} & \rotatebox[origin=c]{90}{Sunglasses} &\rotatebox[origin=c]{90}{Hat} & \rotatebox[origin=c]{90}{T-shirt} & \rotatebox[origin=c]{90}{Long sleeve} & \rotatebox[origin=c]{90}{Formal} & \rotatebox[origin=c]{90}{Shorts} & \rotatebox[origin=c]{90}{Jeans} & \rotatebox[origin=c]{90}{Long Pants} & \rotatebox[origin=c]{90}{Skirt} & \rotatebox[origin=c]{90}{Face Mask} & \rotatebox[origin=c]{90}{Logo} & \rotatebox[origin=c]{90}{Plaid} & \textbf{mAP} \\
		\midrule
		\textbf{Imbalance Ratio} & 1:1 & 1:3 & 1:18 & 1:3 & 1:4 & 1:1 & 1:13 & 1:6 & 1:11 & 1:2 & 1:9 & 1:28 & 1:3 & 1:18 &  \\
		\midrule
		RCNN~\cite{girshick2015fast} & 94 & 81 & 60 & 91 & 76 & 94 & 78 & 89 & 68 & 96 & 80 & 72 & 87 & 55 & 80.0 \\
		R*CNN~\cite{rstarcnn} & 94 & 82 & 62 & 91 & 76 & 95 & 79 & 89 & 68 & 96 & 80 & 73 & 87 & 56 & 80.5 \\
		DHC~\cite{li2016human} & 94 & 82 & 64 & 92 & 78 & 95 & 80 & 90 & 69 & 96 & 81 & 76 & 88 & 55 & 81.3 \\
		VeSPA~\cite{sarfraz2017deep} & - & - & - & - & - & - & - & - & - & - & - & - & - & - & 82.4 \\
		CAM~\cite{guo2017human} & 95 & 85 & 71 & \textbf{94} & 78 & 96 & 81 & 89 & 75 & 96 & 81 & 73 & 88 & 60 & 82.9 \\
		ResNet-101~\cite{resNets2016} & 94 & 85 & 69 & 91 & 80 & 96 & 83 & 91 & 78 & 95 & 82 & 74 & 89 & 65 & 83.7 \\
		ResNet-101+MTL & 94 & 86 & 68 & 91 & 81 & \textbf{96} & 83 & 91 & 79 & 95 & 83 & 74 & \textbf{90} & 65 & 83.8 \\
		ResNet-101+MTL+CRL~\cite{dong2017class} & 94 & 86 & 71 & 91 & 81 & \textbf{96} & 83 & 92 & 79 & 96 & 84 & 76 & \textbf{90} & 66 & 84.7 \\
		SRN~\cite{zhu2017learning}* & 95 & 87 & 72 & 92 & 82 & 95 & 84 & 92 & 80 & \textbf{96} & 84 & 76 & \textbf{90} & 66 & 85.1 \\
		\midrule
		\textbf{Ours} & \textbf{96} & \textbf{88} & \textbf{74} & 93 & \textbf{83} & \textbf{96} & \textbf{85} & \textbf{93} & \textbf{81} & \textbf{96} & \textbf{85} & \textbf{78} & \textbf{90} & \textbf{68} & \textbf{86.4}\\
		\bottomrule
	\end{tabular}
\end{table}

\subsection{Results on WIDER-Attribute}
\noindent~\textbf{Dataset Description and Evaluation Metrics}: The WIDER-Attribute~\cite{li2016human} dataset contains 13,789 images with 57,524 bounding boxes of humans with 14 binary attribute annotations each. Besides ``gender'', which is balanced, the rest of the attributes demonstrate class imbalance, which can reach \(1:18\) and \(1:28\) for attributes such as ``face-mask'' and ``sunglasses''. Following the training protocol of \cite{sarfraz2017deep,zhu2017learning}, we used the human bounding box as an input to our model and mean average precision (mAP) results are reported. 

\noindent~\textbf{Baselines}: We evaluate our approach against all the methods that have been tested on the WIDER-Attribute dataset, namely R-CNN~\cite{girshick2015fast}, R*CNN~\cite{rstarcnn}, DHC~\cite{li2016human}, CAM~\cite{guo2017human}, VeSPA~\cite{sarfraz2017deep}, SRN~\cite{zhu2017learning}, and a fine-tuned ResNet-101 network~\cite{resNets2016}. In addition, we transform the last part of the network to perform multi-task classification (MTL) by adding a fully-connected layer with 64 units for each attribute. This enables us to additionally evaluate against CRL~\cite{dong2017class} by forming triplets within the batch using class-level hard samples. Note that DHC and R*CNN leverage additional contextual information (\eg, scene context or image parts) that intuitively should boost the performance and VeSPA, which jointly predicts the viewpoint along with the attributes, did not train its viewpoint prediction sub-network on the WIDER-Attribute dataset. In SRN~\cite{zhu2017learning}, the validation set was included in the training (which results in \(20\%\) more training data) and samples from the test set were used to obtain an idea about the training performance. In order to allow for a fair comparison with the rest of the methods, we re-implemented their method (which is why there is an asterisk next to their work in Table~\ref{tab:WIDER}) and trained it only on the training set of the WIDER-Attribute~\cite{li2016human} dataset. The difference between the reported results and our re-implementation is \(1.2\) in terms of mAP which is reasonable given the access to approximately \(20\%\) less training data. 

\noindent~\textbf{Evaluation Results}: Our proposed approach achieves state-of-the-art results on the WIDER dataset by improving upon the second best work by 1.3 in terms of mAP and by 2.7 over ResNet-101~\cite{resNets2016} which was our primary network. The larger improvements achieved by our algorithm are in imbalanced attributes such as ``Sunglasses'' or ``Plaid'' that have visual cues in the image which demonstrates the importance of handling class imbalance and using visual attention to identify important visual information in the image. DHC and R*CNN that use additional context information performed significantly worse but this is partially because they utilize smaller primary networks. Overall the proposed approach performs better than or equal than the rest of the literature in all but one attributes and comes second behind CAM~\cite{guo2017human} at recognizing hats.

\setlength{\tabcolsep}{.04cm}
\begin{table}[t]
	\centering
	\caption{Ablation studies on the WIDER dataset to assess the impact of individual modules on the final performance of our method. On the left, we report mAP results just for the primary network (w/o adding any attention mechanisms) using different backbone architectures. On the right, we investigate the additions in terms of performance for attention at a single- and multi-scale level as well as the two loss functions we introduced.}
	\label{tab:wider-abl}
	\footnotesize
	\begin{tabular}[t]{lr|c}
		\toprule
		Primary Net & \textbf{Params} & \textbf{mAP} \\
		\midrule
		ResNet-50 & 25.6\small $\times 10^6$ & 82.3 \\
		DenseNet-121 & 8.1\small$\times 10^6$ & 82.9\\
		ResNet-101 &44.7\small$\times 10^6$ & 83.7\\
		ResNet-152  &60.4\small$\times 10^6$ & 84.2\\
		DenseNet-201 &20.2\small$\times 10^6$ & 84.5\\
		\bottomrule
	\end{tabular}
	\; \;
	\begin{tabular}[t]{c|cccc|c}
		\toprule
		Primary Net & \(\mathcal{L}_{w}\) & Attention & \(\mathcal{L}_{a} \) & Multi-scale & \textbf{mAP} \\
		\midrule
		ResNet-101 & & & & & 83.7 \\
		ResNet-101 & \checkmark & & & & 84.4 \\
		ResNet-101 & \checkmark & \checkmark & & & 85.0 \\
		ResNet-101 & \checkmark & \checkmark & \checkmark &  & 85.7 \\
		ResNet-101 & \checkmark & \checkmark &  & \checkmark & 85.9 \\
		ResNet-101 & \checkmark & \checkmark & \checkmark & \checkmark & 86.4 \\
		\bottomrule
	\end{tabular}
\end{table} 

\subsection{Ablation Studies on WIDER}
In our first ablation study (Table~\ref{tab:wider-abl} - left), we investigate to what extent the primary network affects the final performance. This is because it is commonplace that as architectures become deeper, the impact of individual add-on modules becomes less significant. We observe that (i) the difference between a ResNet-50 and a DesneNet-201 architecture is more than \(2\%\) in terms of mAP, (ii) DenseNet-201, which is the highest performing primary network, is almost as good as SRN~\cite{zhu2017learning} due to its effective feature aggregation and reuse, and (iii) the mAP of the proposed approach is \(2.1\) more than the best performing primary network. In our second ablation study (Table~\ref{tab:wider-abl} - right), we assess how each proposed component of our approach contributes to the final mAP. Our ResNet-101 baseline (w/o any class weighting) achieves 83.7\% mAP which increases to 84.0\% when the class weights are added. When the instance-level weighting is added (\ie, \(L_{w}\)) the total performance increases to 84.4\%. These results indicate that it is important to take both class-level and instance-level weighting into consideration during imbalanced learning. Handling class imbalance using the weighted focal loss and adding our attention mechanism just at a single scale result in mAP equal to \(85.0\) which performs almost as well as the existing state-of-the-art. Adding the attention loss that penalizes attention masks with high prediction variance and expanding our attention module to two scales improves the final mAP to 86.4.

\begin{figure*}[t]
	\centering
	\includegraphics[width=0.95\linewidth]{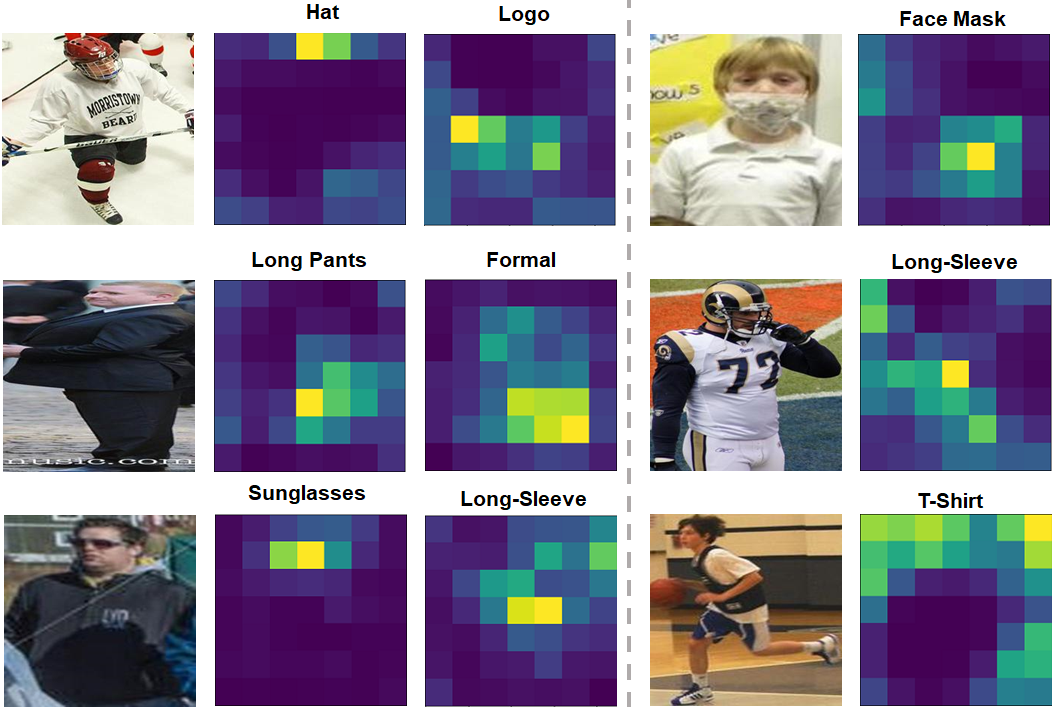}
	\caption{Successful attention masks (left) and failure cases (right) for attributes of the WIDER dataset. The attention masks of our method try to find formal clothes and long pants in the bottom part of the image, logos in the middle and sunglasses or hats at the top. However, due to their weakly-supervised training, there are examples in which they fail to identify the correct locations (face mask in the bottom) or make completely wrong guesses as in the T-shirt example in the bottom right. }
	\label{fig:attentionMasks}
\end{figure*}

\noindent~\textbf{Qualitative Results}:	Figure~\ref{fig:attentionMasks}, depicts attention masks for six successful (left) and three failure cases (right). We observe that for imbalanced attributes such as sunglasses that have discriminant visual cues, our attention mechanism locates successfully the corresponding regions, which explains the \(7\%\) relative improved mAP for this attribute compared to our primary ResNet architecture. 

\setlength{\tabcolsep}{.35cm}
\begin{table}[t]
	\centering
	\caption{Evaluation of the proposed approach against nine different approaches on the PETA dataset ranked by F1-score. The asterisk next to SRN indicates that it is our re-implementation due to the fact that the validation set was included in the original work which is not the case for the rest of the methods. The loss next to it corresponds to the loss function used in each case.}
	\label{tab:peta_eval}
	\begin{tabular}[t]{l ccccc}
		\toprule
		\textbf{Method} & \textbf{mA} & \textbf{Acc} & \textbf{Prec} & \textbf{Rec} & \textbf{F1} \\
		\midrule
		ACN~\cite{sudowe2015person} & 81.15 & 73.66 & 84.06 & 81.26 & 82.64 \\
		SRN~\cite{zhu2017learning}* (w/ \(\mathcal{L}_b\)) & 80.55 & 74.24 & 84.04 & 82.48 &  83.25\\
		WPAL-FSPP~\cite{yu2016weakly} & 84.16 & 74.62 & 82.66 & 85.16 & 83.40 \\
		DeepMAR~\cite{li2015multi} & 82.89 & 75.07 & 83.68 & 83.14 & 83.41 \\
		GoogleNet~\cite{szegedy2015going} & 81.98 & 76.06 & 84.78 & 83.97 & 84.37 \\
		ResNet-101~\cite{resNets2016} & 82.67 & 76.63 & 85.13 & 84.46 & 84.79 \\
		WPAL-GMP~\cite{yu2016weakly} & \textbf{85.50} & 76.98 & 84.07 & 85.78 & 84.90 \\
		SRN~\cite{zhu2017learning}* (w/ \(\mathcal{L}_{w}\)) & 82.36 & 75.69 & 85.25 & 84.59 & 84.92 \\
		VeSPA~\cite{sarfraz2017deep} & 83.45 & 77.73 & 86.18 & 84.81 & 85.49 \\
		\midrule
		\textbf{Ours} & 84.59  & \textbf{78.56} & \textbf{86.79} & \textbf{86.12} & \textbf{86.46} \\
		\bottomrule
	\end{tabular}
\end{table}

\subsection{Results on PETA}
\noindent~\textbf{Dataset Description and Evaluation Metrics}: The PETA~\cite{deng2014pedestrian} dataset is a collection of 10 person surveillance datasets and consists of 19,000 cropped images along with 61 binary and 5 multi-value attributes. We used the same train/validation/test splits with the method of Sarfraz \etal~\cite{sarfraz2017deep} and followed the established protocol of this dataset by reporting results on the 35 attributes for which the ratio of positive labels is higher than 5\%. For the PETA dataset, two different types of metrics are reported namely label-based and example-based~\cite{li2016richly}. For the label-based metrics due to the imbalanced class distribution, we used the balanced mean accuracy (mA) for each attribute that computes separately the classification accuracy of the positive and the negative examples  and then computes the average. For the label-based metrics, we report accuracy, precision, recall, and F1-score averaged across all examples in the test set. 

\noindent~\textbf{Baselines}: We compared our approach with all the methods that have been tested on the PETA dataset, namely the ACN~\cite{sudowe2015person}, DeepMAR~\cite{li2015multi}, two variations of WPAL~\cite{yu2016weakly}, VeSPA~\cite{sarfraz2017deep}, the GoogleNet~\cite{szegedy2015going} baseline reported by Sarfraz \etal~\cite{sarfraz2017deep}, ResNet-101~\cite{resNets2016} and SRN~\cite{zhu2017learning}. 

\noindent~\textbf{Evaluation Results}: From the complete evaluation results in Table~\ref{tab:peta_eval}, we observe that the proposed approach achieves state-of-the-art results in all example-based metrics and comes second to WPAL~\cite{yu2016weakly} in terms of balanced mean accuracy (mA). We believe this is due to the fact that different methods use different metrics, based on which they optimize their models. For example, our approach is optimized based on the F1 score which balances between precision and recall and is applicable in search applications. Our approach improves upon a fine-tuned ResNet-101 architecture by approximately \(2\%\) in terms of F1 score which demonstrates the importance of the visual attention mechanisms. Notably, we improve upon VeSPA~\cite{sarfraz2017deep} in all evaluation metrics despite the fact that they utilize additional viewpoint information to train their model. Finally, we observe that by using the weighted variant of focal loss (\(\mathcal{L}_{w}\)) instead of the binary-cross entropy loss (\(\mathcal{L}_b\)), the F1 score of SRN~\cite{zhu2017learning} increases by \(1.7\%\). This demonstrates why failing to account for class imbalance affects the performance of deep attribute classification models. 

\subsection{Ablation Studies on PETA} Based on our analysis an important question arises: can we achieve similar results with significantly fewer parameters? Aiming to find out the impact of large backbone architectures in the final performance, we investigated how each component of our work performs using a pre-trained DenseNet-121~\cite{huang2017densely} architecture. DenseNet-121 contains \(7.5 \times \) less parameters compared to ResNet-101 due to efficient feature propagation and reuse. To our surprise, when all components are included (last row in Table~\ref{tab:peta_abl}), the performance drop in terms of F1 score is less than \(2\%\). In addition, we explored a variety of feature aggregations by either up-sampling the smaller attention masks, max-pooling the larger or mapping the larger to the smaller using a convolutional layer with a stride equal to two. Although the latter approach performed better than up-sampling/down-sampling, we observed that the aggregation of the attention information at a logit level is superior compared to  feature level aggregation. We believe that this is because the two attention mechanisms extract masks that give emphasis to different spatial regions that when added together fail to provide the classifier with attribute-discriminative information.

\setlength{\tabcolsep}{.1cm}
\begin{table}[t]
	\centering
	\caption{Ablation studies to assess the impact of each submodule to the final result using DenseNet-121 as a light-weight backbone architecture.}
	\label{tab:peta_abl}
	\small    
	\begin{tabular}[t]{c|P{1cm}cc P{2.1cm} P{2.1cm} |c}
		\toprule
		Primary Net & Class Weight & \(\mathcal{L}_{w}\) & Attention & Multi-scale (feature aggr.) &  Multi-scale (score aggr.)& \textbf{F1} \\
		\midrule
		DenseNet-121 & \checkmark & & & & &  82.1 \\
		DenseNet-121 & \checkmark & \checkmark & & & & 82.9 \\
		DenseNet-121 & \checkmark & \checkmark & \checkmark & & & 83.8 \\
		DenseNet-121 & \checkmark & \checkmark & \checkmark & \checkmark & & 84.1 \\
		DenseNet-121 & \checkmark & \checkmark & \checkmark &  & \checkmark & 84.7 \\
		\bottomrule
	\end{tabular}
\end{table}

\subsection{Sources of Error and Further Improvements}
\begin{figure*}[t]
	\centering
	\includegraphics[width=0.99\linewidth]{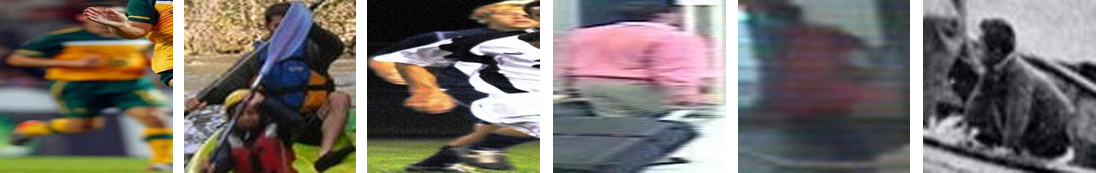}
	\caption{Pedestrian attribute datasets contain images with large inherent noise and variation. Images can be out of focus, occluded, wrongly cropped, resized to fixed squared higher resolutions, blurry or even grayscale.}
	\label{fig:problematic}
\end{figure*}

Where does the proposed method fail and what are the characteristics of the failure cases? Aiming to gain a better understanding we will discuss separately the errors originating from the noise inherent to the input data and the errors related to modeling. A significant limitation of most pedestrian attribute classification methods (including ours) is that they resize the input data to a fixed square-size resolution (\eg, \(224\times224\)) in order to feed them to deep pre-trained architectures. Human crops are usually rectangular captured from different viewpoints and thus, when they are resized to a square, important spatial information is lost. One possible solution to this would be feeding the whole image (before performing the human crop) at a fixed resolution that does not destroy the spatial relations and then extract human-related features using ROI-pooling at a stage within the network. To cope with the high viewpoint variance, the spatial transformer networks of Jaderberg \etal~\cite{jaderberg2015spatial} could be employed to align the input image before feeding it to the network, a practice which is very common in face recognition applications~\cite{peng2016recurrent,tuzel2016robust,jeni2017dense}. A second source of error is the very low resolution of several images especially in the PETA dataset, which makes it hard even for the human eye to identify the attribute traits of the depicted human. Some training examples that demonstrate these sources of error are depicted in Figure~\ref{fig:problematic}. In addition, the provided annotations contain a third unspecified/uncertain class, which is used as negative during training in the literature, that further dilutes the learning process. Applying modern super-resolution techniques \cite{kim2016accurate,dahl2017pixel} could alleviate this issue but only to some extent. Regarding errors due to modeling richer feature representations could be extracted using feature pyramid networks~\cite{lin2017feature} since they extract high-level semantic feature maps at multiple scales. Because the goal of this paper was to introduce a simple yet effective attribute classification solution, we refrained from building a complicated attention mechanism with a high number of parameters. Modern visual attention mechanisms~\cite{li2018harmonious,wang2017residual,liang2018focal} could be adapted to a multi-label setup and applied to achieve superior performance at the expense of a larger parameter space. 

\section{Conclusion}
Learning the visual attributes of humans is a multi-label classification problem that suffers from large class imbalance and lack of semantic/spatial attribute annotations. To address these challenges, we developed a simple yet effective and easy-to-reproduce architecture that outputs visual attention masks at multiple scales and handles effectively class imbalance and samples with high prediction variance. We introduced a weighted variant of focal loss that handles the prior class imbalance per attribute and focuses on hard misclassified positive samples. In addition, we observed that the weakly-supervised attention masks result in high prediction variance and thus, we introduced an attention loss that penalizes accordingly such predictions. By simplifying the problem and addressing each one of its challenges, we achieve state-of-the-art results in both the WIDER-Attribute and PETA datasets, which are the most widely used in this domain. This work aspires to serve as a bar in the visual attribute classification domain that future works can improve upon. To facilitate this process, we performed ablation studies, identified some sources of error that still exist and pointed out possible future research directions that require further exploration. 
\linebreak

\noindent \textbf{Acknowledgments}: This work has been funded in part by the UH Hugh Roy and Lillie Cranz Cullen Endowment Fund. All statements of fact, opinion or conclusions contained herein are those of the authors and should not be construed as representing the official views or policies of the sponsors.

\bibliographystyle{splncs}
\bibliography{Refs}
\clearpage
\section*{Supplementary Material}
\subsection*{Training Details}
Since in both datasets we used a pre-trained primary network we first froze its weights and learned the attention masks using their corresponding loss function. This was done, to avoid back-propagating large prediction errors from the attention masks to the pre-trained network. After a few epochs of training solely the attention mechanism, the primary network is then unfrozen and trained end-to-end to produce multi-attribute predictions. For the WIDER-Attribute dataset we set the learning rate equal to \(0.001\) and use SGD with momentum set to \(0.9\) and a weight decay equal to \(0.0005\). The learning rate was divided by \(10\) (until \(0.00001\)) when the error plateaus in the validation set. During pre-processing, we resized all images to \(256\times 256\) and extracted random crops of \([128,224]\) (along with random mirroring and data shuffling) which were then resized to \(224\times 224\) and provided as an input to the network. For the PETA dataset we used Adam since it consistently outperformed SGD with a starting learning rate equal to \(0.0001\) with the same weight decay but with larger crops (in the range \([160,224]\)). In both datasets, the batch size was set to \(32\). We used MXNet/Gluon as our deep learning framework and a single NVIDIA GeForce GTX 1080 Ti GPU.

\subsection*{Architecture Details}
Our backbone architecture is a ResNet-101 that extracts feature representations of dimensionality \(7\times 7 \times 2048\) which are then fed to a fully-connected layer. Its dimensionality is equal to the number of classes denoted by \(\mathcal{C}_l\) which for the WIDER dataset is equal to 14. The attention modules are placed on ``stage3\_activation22'' and ``stage4\_activation2''. Let Ck denote a Convolution-BatchNorm-ReLU layer
with k filters and kernel size equal to 1 and Dk a fully-connected layer with k neurons. The attention module consists of C256-C256 and a convolutional layer with \(\mathcal{C}_l\) number of filters. Its output is first spatially normalized and then multiplied by the output of the confidence weighting layer which is simply a convolutional layer with \(\mathcal{C}_l\) number of filters and a sigmoid activation function. The output of the attention modules is fed to a C256-C512-C512-D\(\mathcal{C}_l\) subnetwork the last convolutional layer of which has a kernel size equal to the spatial dimensions. All layers are initialized with Xavier initialization.
\end{document}